\documentclass[9pt,conference]{IEEEtran}
\usepackage{amssymb,amsthm,amsmath,array}
\usepackage{graphicx}
\usepackage[caption=false,font=footnotesize]{subfig}
\usepackage{xspace}
\usepackage[sort&compress, numbers]{natbib}
\usepackage{stmaryrd}
\usepackage{xcolor}
\usepackage{mathtools}
\usepackage{float}
\usepackage{textcomp}
\usepackage{comment}
\usepackage{multirow}
\usepackage{color}

\begin{document}
\title{Impact of Disentanglement on Pruning Neural Networks}
\author{\IEEEauthorblockN{
        Carl Shneider\IEEEauthorrefmark{1},
        Peyman Rostami\IEEEauthorrefmark{1}, 
        Anis Kacem\IEEEauthorrefmark{1},
        Nilotpal Sinha\IEEEauthorrefmark{1}, 
        Abd El Rahman Shabayek\IEEEauthorrefmark{1}, and 
        Djamila Aouada\IEEEauthorrefmark{1}       
    }
    \IEEEauthorblockA{
        \IEEEauthorrefmark{1}Interdisciplinary Centre for Security, Reliability and Trust, University of Luxembourg, Luxembourg
       }
}
\maketitle
\begin{abstract}
Deploying deep learning neural networks on edge devices, to accomplish task specific objectives in the real-world, requires a reduction in their memory footprint, power consumption, and latency. 
This can be realized via efficient model compression. 
Disentangled latent representations produced by variational autoencoder (VAE) networks are a promising approach for achieving model compression because they mainly retain task-specific information, discarding useless information for the task at hand.   
We make use of the Beta-VAE framework combined  with a standard criterion for pruning to investigate the impact of forcing the network to learn disentangled representations on the pruning process for the task of classification. 
In particular, we perform experiments on MNIST and CIFAR10 datasets, examine disentanglement challenges, and propose a path forward for future works.

\end{abstract}


\section{Introduction}

Advances in deep learning have accelerated the state-of-the-art in computational sensing across research domains, finding wide application in tasks encompassing object detection \cite{Bochkovskiy_YOLOv4, perez2021detection}, pose estimation \cite{lauer2022multi, musallam2022leveraging, SurveySPE23}, classification \cite{singh2022multi}, reconstruction \cite{CIRIM22}, segmentation \cite{Li_2022_CVPR, Cheng_2022_CVPR}, time series prediction \cite{borovykh2017conditional, mejri2022unsupervised}, and 3D modelling \cite{Dupont3DV22, karadeniz2023tscom}.
Surpassing a top performing deep neural network (DNN) on image-level and pixel-level tasks on a benchmark dataset often comes at the cost of an increased number of model parameters, floating point operations (FLOPS), volume of training data, and graphical processing unit (GPU) resources.
This trend poses a challenge for the deployment of these pre-trained DL models for real-time inference on resource constrained edge devices, with low computing power and memory, where a reduction in the model's memory footprint (i.e., RAM and storage requirements), power consumption, and latency, is essential. 
With the objective of reducing the model size of an existing pre-trained base model,
while maintaining a comparable level of metric performance (e.g., accuracy, precision, recall, etc.) to that of the original model, different techniques of neural network compression have been proposed. 
These include quantization \cite{BinaryConnect15}, knowledge distillation \cite{KnowlDistil15}, low-rank matrix factorization \cite{LinStruct14}, and pruning \cite{PruningReed93}. 

Quantization compresses the original network by reducing the number of bits required to represent each weight \cite{SurveyCompressHardware}.
This reduces the range of data values while retaining most of the information for comparable model accuracy but under certain conditions 
can lead to a drop in performance.
In knowledge distillation, the knowledge from a pre-trained, larger and more complex model, known as the teacher model, is transferred to a smaller network, known as the student network.
Limitations of this approach arise from the side of the student network whose architecture needs to be designed and usually remains fixed. 
Low-rank factorization can be applied to both convolutional and fully connected layers. Applying tensor decomposition to the convolutional layers makes the inference process faster while the factorization of the matrices in the dense layer makes the model's memory footprint lighter.
A limitation of this approach is that the proper factorization and rank selection process is challenging to implement in practice and is computationally more intensive.
Pruning is used for network sparsification by removing redundant and non-essential parameters, contributing to reduced model size (i.e., smaller memory footprint) and improved latency (i.e., lower bandwidth). 
Pruning can produce either unstructured or structured network sparsification patterns. 
Unstructured pruning involves setting weight values to zero, as in convolutional parameters in a convolutional neural network (CNN) or connections between neurons in a fully convolutional neural network (FCNN). 
A drawback of unstructured pruning is that most machine learning frameworks and hardware cannot accelerate sparse matrices’ computation. 
Structured pruning includes kernel/channel pruning, filter pruning (i.e., neurons in FCNNs), and layer pruning. 
Although a more structured sparsification pattern leads to an accelerated neural network, the accuracy drops \cite{SurveyCompressHardware}. 
Pruning has proven successful at maintaining the level of performance of the original pre-trained network while discarding superfluous information contained in the specified network structures described above. 
Nevertheless, given a specific task, the challenge of selecting which parts of the network to prune remains open, with several criteria proposed to rank the relative importance of the respective components.
These criteria include pruning the weight magnitude or the gradient magnitude, and whether to prune globally or locally.

Neural network compression can be assisted by disentangling latent representations because a representation which is disentangled for a particular dataset is one which is sparse over the transformations present in that data \cite{Bengio13_Whitney16}. 
Disentangled representations are representations that capture the underlying factors of variation that generated the data, especially those explanatory factors that are relevant to the task.
Disentangling as many factors as possible while discarding as little information about the data as is practical for the task at hand, is a robust approach to feature learning \cite{Goodfellow-et-al-2016}. 
In this work, instead of proposing a new pruning criterion, we use a standard low-magnitude criterion for local, unstructured pruning and investigate the impact of enforcing the network to learn disentangled representations on the pruning process. 
The intuition here is that if the network is learning disentangled representations then this would be reflected by the neural network itself which would make pruning more effective. 
Specifically, we make use of the Beta-VAE \cite{Higgins2016betaVAELB}  framework combined with the aforementioned standard criterion for pruning to investigate the impact of forcing the network to learn disentangled representations on the pruning process for the task of classification. 
Our contributions are two-fold: 
\begin{itemize}
    \item we provide preliminary analysis on the impact of disentanglement on pruning neural networks for the task of classification,
    \item we perform experiments on MNIST and CIFAR10 datasets for this objective.
\end{itemize}

\begin{figure}
\centering
\includegraphics[width=0.8\linewidth]{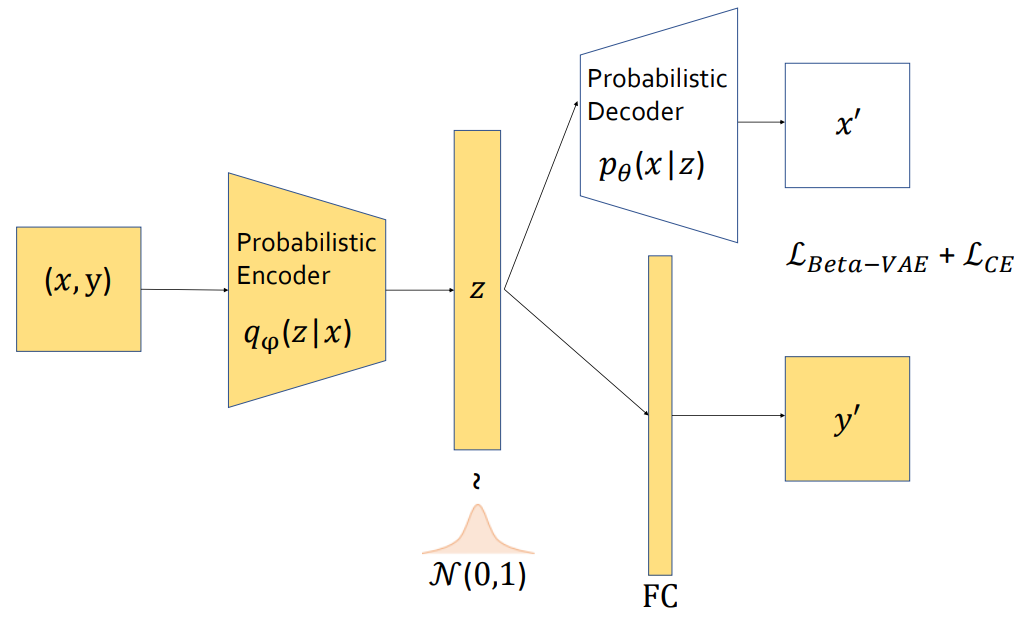}
\label{fig:BVAE}
\caption{The Beta-VAE model augmented by the addition of a classifier head. The combined model, Beta-VAE-Classif, is trained with all three loss terms given by the KL divergence, reconstruction loss, and classification loss. During inference, the reconstruction head is removed, leaving the shaded in blocks of the diagram. The notations used are described in Section~\ref{sec:method}.}
\end{figure}

\section{Proposed Method} \label{sec:method}

Consider a dataset $\mathcal{D} = \{(x_i, y_i)\}^{N}_{i=1}$, $x$ is the 2D input data 
and $y$ is the corresponding label. 
We use the Beta-VAE framework to enforce a disentangled latent representation \cite{Higgins2016betaVAELB} given by

 \begin{equation} \label{eq:bvae}
 \begin{split}
 \mathcal{L_\text{Beta-VAE}}(\theta, \phi; x, z, \beta) = &-\beta~\mathcal{D}_{KL}\left( q_\phi\left(z|x\right)\parallel p\left(z\right)\right)  \\ 
&+ \mathop{{}\mathbb{E}_{q_{\phi}\left(Z|x\right)}}[\log p_\theta\left(x|z\right)]
 \end{split}
\end{equation}
where $p(z)$ is a prior distribution of the latent variables $z$, $p_\theta(x|z)$ is a conditional probability of $x$ that is parametrized by a neural network $\theta$, and $q_\phi(z|x)$ approximates the posterior with neural network parameters $\phi$.

\begin{equation} \label{eq:ce}
 \mathcal{L_\text{CE}}(y) = -\sum_{i=1}^{N}y_i\log y'_i
\end{equation}
\vspace{-0.5em}
\begin{equation} \label{eq:bvae_classif}
\mathcal{L_\text{Beta-VAE-Classif}}(\theta, \phi; x, y, z, \beta) = \mathcal{L_\text{Beta-VAE}} + \mathcal{L_\text{CE}}
\end{equation}

The Beta-VAE objective in Eq.~\ref{eq:bvae} consists of a Kullback-Leibler (KL) divergence term (encoder part) and a pixel-wise reconstruction loss (decoder part). The penalty $\beta$ imposes a limit on the capacity of latent information and the degree to which statistically independent latent factors are learned. 
Our training strategy, 
first involves appending 
a classifier head to the Beta-VAE during training as shown in Fig.~1~
and augmenting the loss function in Eq.~\ref{eq:bvae} with the addition of sparse categorical cross-entropy loss term given by Eq.~\ref{eq:ce} in Eq.~\ref{eq:bvae_classif}.
In Eq.~\ref{eq:ce}, $y'$ is the predicted label.
During inference, the reconstruction head is removed, as well as its corresponding loss term, leaving the pre-trained encoder and classification head.

\begin{table}[h!]
\caption{Accuracy and compression for MNIST and CIFAR10 datasets. Pure CNN models are denoted as CNN-Classif. The augmented Beta-VAE models with the classifier head are denoted as Beta-VAE-Classif. The mean and standard deviation is computed over three independent runs where available. Compression is given in terms of kilobytes for MNIST and megabytes for CIFAR10.}
\begin{tabular}{ |l|c|l|l|  } 
 \hline
 \multicolumn{4}{|c|}{MNIST} \\
 \hline
 \hline
 Model & Beta & Acc. (\%) & Comp. (KB) \\
 \hline
 CNN-Classif.                       & - & 98.19 $\pm$ 0.2 & 260.28 $\pm$ 0.1  \\
 CNN-Classif. w/ pruning            & - & 97.29 $\pm$ 0.4 & 84.25 $\pm$ 0.3 \\
  \hline
 Beta-VAE Classif.              & 1 & 96.32 $\pm$ 0.5 & 260.89 $\pm$ 0.0 \\
 Beta-VAE Classif. w/ pruning   & 1 & 90.02 $\pm$ 7.3 & 84.43 $\pm$ 0.2 \\
 \hline
 Beta-VAE Classif.              & 3 & 95.34 $\pm$ 1.7 & 260.78 $\pm$ 0.6 \\
 Beta-VAE Classif. w/ pruning   & 3 & 92.52 $\pm$ 2.5 & 84.81 $\pm$ 0.4   \\
 \hline
 Beta-VAE Classif.              & 5 & 96.86 $\pm$ 1.0 & 261.37 $\pm$ 0.0 \\
 Beta-VAE Classif. w/ pruning   & 5 & 96.15 $\pm$ 0.2 & 85.29 $\pm$ 0.2 \\
 \hline
 Beta-VAE Classif.              & 10 & 95.67 $\pm$ 0.6 & 260.43 $\pm $ 0.1 \\
 Beta-VAE Classif. w/ pruning   & 10 & 95.86 $\pm$ 0.8 & 85.67 $\pm$ 0.4  \\
 \hline
 \multicolumn{4}{|c|}{CIFAR10} \\
 \hline
 \hline
 Model & Beta & Acc. (\%) & Comp. (MB) \\
 \hline
 CNN-Classif.                       & -  & 78.14 $\pm$ 0.0  & 10.9 $\pm$ 0.000 \\
 CNN-Classif. w/ pruning            & -  & 78.12 $\pm$ 0.0 & 3.5 $\pm$ 0.000   \\
  \hline
 Beta-VAE-Classif.              & 1  & 48.97 $\pm$ 1.4 & 10.83 $\pm$ 0.002  \\
 Beta-VAE-Classif. w/ pruning   & 1  & 48.76 $\pm$ 5.2 & 3.43 $\pm$ 0.005 \\
 \hline
 Beta-VAE-Classif.              & 3  & 51.08  $\pm$ 4.5 & 10.82 $\pm$ 0.001 \\
 Beta-VAE-Classif. w/ pruning   & 3  & 49.30 $\pm$ 5.9 & 3.44 $\pm$ 0.009 \\
 \hline
 Beta-VAE-Classif.              & 5  & 47.17 $\pm$ 9.4 & 10.82 $\pm$ 0.001 \\
 Beta-VAE-Classif. w/ pruning   & 5  & 51.28 $\pm$ 10.2 & 3.43 $\pm$ 0.009 \\
 \hline
 Beta-VAE-Classif.              & 10 & 10.00 $\pm$ 0.0  & 10.80 $\pm$ 0.004 \\
 Beta-VAE-Classif. w/ pruning   & 10 & 10.00 $\pm$ 0.0  & 3.35 $\pm$ 0.001 \\
 \hline
\end{tabular}
\label{cap:tab1}
\end{table}
\vspace{-0.5em}

\section{Experiments and Discussions} \label{sec:results}

Our experiments are implemented in the TensorFlow framework using MNIST and CIFAR10 datasets.
The MNIST dataset consists of 70k 28x28 grayscale images of the 10 digits of which 60k are used for the training set and 10k for the test set. The CIFAR10 dataset consists of 60k 32x32 color images, labeled across 10 categories, of which 50k are used for training and 10k for testing. 
The Adam optimizer is used with a learning rate of $0.001$ and data are shuffled at each epoch during training. 
The Beta parameter assumes values of $\beta = \{1,3,5,10\}$ and experiments are performed with and without pruning for each of these $\beta$ values.
Initially all models are trained for 30 epochs with 2 further epochs once the reconstruction head is removed. 
At this stage, if pruning is applied, an additional 2 epochs are used for this purpose followed by the evaluation of the classification accuracy. 
We use the `prune\_low\_magnitude' function in Keras with a 50\% constant sparsity which is a standard low-magnitude criterion for local, unstructured pruning.
The neural network architectures used for the MNIST and CIFAR10 datasets are parameterized by 69k and 2.9M parameters, respectively.

The MNIST dataset is a far easier dataset than CIFAR10 and, consequently, easier to over-train on as seen from the high accuracies attained both with the pure CNN model and Beta-VAE-Classif model up to $\beta=10$ in Table~\ref{cap:tab1}.
The pure CNN-Classif model baseline outperforms the Beta-VAE-Classif models on both datasets because of its singular objective compared to having to optimize multiple objectives coming from several tasks.
Preliminary results of several individual runs on MNIST initially hinted that at a certain degree of disentanglement $\beta > 1$ the accuracy of the pruned and unpruned models increased over the $\beta = 1$ entanglement baseline. 
This finding would have supported the intuition that applying disentanglement to a model initially guided by the presence of a pixel-level task, such as reconstruction, would facilitate discarding information that is not relevant for the relatively simpler, image-level task of classification. 
This observation may be captured by the $\beta=5$ Beta-VAE-Classif model on the CIFAR10 dataset results since the upper range of accuracy exceeds that of the $\beta=1$ model. 
The compression ratio of pruned to unpruned model are essentially the same among all models examined for each of the two datasets.  
The severe drop in accuracy for the $\beta=10$ on the CIFAR10 dataset appear to be indicative of the information preference problem (i.e., posterior collapse, KL vanishing) where the disentangled representations have become independent of the observation space. 

\vspace{-0.5em}
\section{Conclusion and Future Works} \label{sec:concl}
\vspace{-0.5em}

The preliminary results presented in this work are inconclusive and only roughly suggest that for a certain value of $\beta$ in the Beta-VAE framework, the latent space implicitly becomes sufficiently disentangled to allow for pruning to more easily discard useless information for the task of classification. 
Increasing the number of epochs over which pruning is performed as well as the number of latent dimensions may improve these results.
Furthermore, having ground truth disentanglement labels together with an appropriately selected metric to directly measure the degree of disentanglement for the task of classification is expected to yield more robust results. We propose to use the dSprites \cite{dsprites17} dataset along with the mutual information gap (MIG) \cite{chen2018isolating} metric for this purpose. 

\section{Acknowledgement}
This work is supported by the Luxembourg National Research Fund (FNR), under the project reference C21/IS/15965298/ELITE.
\bibliographystyle{IEEEtran}
\bibliography{references}

\begin{thebibliography}{10}
\providecommand{\url}[1]{#1}
\csname url@samestyle\endcsname
\providecommand{\newblock}{\relax}
\providecommand{\bibinfo}[2]{#2}
\providecommand{\BIBentrySTDinterwordspacing}{\spaceskip=0pt\relax}
\providecommand{\BIBentryALTinterwordstretchfactor}{4}
\providecommand{\BIBentryALTinterwordspacing}{\spaceskip=\fontdimen2\font plus
\BIBentryALTinterwordstretchfactor\fontdimen3\font minus
  \fontdimen4\font\relax}
\providecommand{\BIBforeignlanguage}[2]{{%
\expandafter\ifx\csname l@#1\endcsname\relax
\typeout{** WARNING: IEEEtran.bst: No hyphenation pattern has been}%
\typeout{** loaded for the language `#1'. Using the pattern for}%
\typeout{** the default language instead.}%
\else
\language=\csname l@#1\endcsname
\fi
#2}}
\providecommand{\BIBdecl}{\relax}
\BIBdecl

\bibitem{Bochkovskiy_YOLOv4}
\BIBentryALTinterwordspacing
A.~Bochkovskiy, C.-Y. Wang, and H.-Y.~M. Liao, ``Yolov4: Optimal speed and
  accuracy of object detection,'' 2020. [Online]. Available:
  \url{https://arxiv.org/abs/2004.10934}
\BIBentrySTDinterwordspacing

\bibitem{perez2021detection}
M.~Perez, M.~A. Mohamed~Ali, A.~Garcia~Sanchez, E.~Ghorbel, K.~Al~Ismaeil,
  P.~Le~Henaff, and D.~Aouada, ``Detection \& identification of on-orbit
  objects using machine learning,'' in \emph{European Conference on Space
  Debris}, vol.~8, no.~1, 2021.

\bibitem{lauer2022multi}
J.~Lauer, M.~Zhou, S.~Ye, W.~Menegas, S.~Schneider, T.~Nath, M.~M. Rahman,
  V.~Di~Santo, D.~Soberanes, G.~Feng \emph{et~al.}, ``Multi-animal pose
  estimation, identification and tracking with deeplabcut,'' \emph{Nature
  Methods}, vol.~19, no.~4, pp. 496--504, 2022.

\bibitem{musallam2022leveraging}
M.~A. Musallam, V.~Gaudilliere, M.~O. Del~Castillo, K.~Al~Ismaeil, and
  D.~Aouada, ``Leveraging equivariant features for absolute pose regression,''
  in \emph{Proceedings of the IEEE/CVF Conference on Computer Vision and
  Pattern Recognition}, 2022, pp. 6876--6886.

\bibitem{SurveySPE23}
L.~Pauly, W.~Rharbaoui, C.~Shneider, A.~Rathinam, V.~Gaudilliere, and
  D.~Aouada, ``A survey on deep-learning based monocular spacecraft pose
  estimation: Algorithms and datasets,'' in prep.

\bibitem{singh2022multi}
I.~P. Singh, E.~Ghorbel, O.~Oyedotun, and D.~Aouada, ``Multi label image
  classification using adaptive graph convolutional networks (ml-agcn),'' in
  \emph{2022 IEEE International Conference on Image Processing (ICIP)}.\hskip
  1em plus 0.5em minus 0.4em\relax IEEE, 2022, pp. 1806--1810.

\bibitem{CIRIM22}
D.~{Karkalousos}, S.~{Noteboom}, H.~E. {Hulst}, F.~M. {Vos}, and M.~W.~A.
  {Caan}, ``{Assessment of data consistency through cascades of independently
  recurrent inference machines for fast and robust accelerated MRI
  reconstruction},'' \emph{Physics in Medicine and Biology}, vol.~67, no.~12,
  p. 124001, Jun. 2022.

\bibitem{Li_2022_CVPR}
L.~Li, T.~Zhou, W.~Wang, J.~Li, and Y.~Yang, ``Deep hierarchical semantic
  segmentation,'' in \emph{Proceedings of the IEEE/CVF Conference on Computer
  Vision and Pattern Recognition (CVPR)}, June 2022, pp. 1246--1257.

\bibitem{Cheng_2022_CVPR}
B.~Cheng, I.~Misra, A.~G. Schwing, A.~Kirillov, and R.~Girdhar,
  ``Masked-attention mask transformer for universal image segmentation,'' in
  \emph{Proceedings of the IEEE/CVF Conference on Computer Vision and Pattern
  Recognition (CVPR)}, June 2022, pp. 1290--1299.

\bibitem{borovykh2017conditional}
A.~Borovykh, S.~Bohte, and C.~W. Oosterlee, ``Conditional time series
  forecasting with convolutional neural networks,'' \emph{arXiv preprint
  arXiv:1703.04691}, 2017.

\bibitem{mejri2022unsupervised}
N.~Mejri, L.~Lopez-Fuentes, K.~Roy, P.~Chernakov, E.~Ghorbel, and D.~Aouada,
  ``Unsupervised anomaly detection in time-series: An extensive evaluation and
  analysis of state-of-the-art methods,'' \emph{arXiv preprint
  arXiv:2212.03637}, 2022.

\bibitem{Dupont3DV22}
E.~Dupont, K.~Cherenkova, A.~Kacem, S.~A. Ali, I.~Aryhannikov, G.~Gusev, and
  D.~Aouada, ``Cadops-net: Jointly learning cad operation types and steps from
  boundary-representations,'' in \emph{2022 International Conference on 3D
  Vision (3DV)}, 2022.

\bibitem{karadeniz2023tscom}
A.~S. Karadeniz, S.~A. Ali, A.~Kacem, E.~Dupont, and D.~Aouada, ``Tscom-net:
  Coarse-to-fine 3d textured shape completion network,'' in \emph{Computer
  Vision--ECCV 2022 Workshops: Tel Aviv, Israel, October 23--27, 2022,
  Proceedings, Part V}.\hskip 1em plus 0.5em minus 0.4em\relax Springer, 2023,
  pp. 289--306.

\bibitem{BinaryConnect15}
M.~{Courbariaux}, Y.~{Bengio}, and J.-P. {David}, ``{BinaryConnect: Training
  Deep Neural Networks with binary weights during propagations},'' \emph{arXiv
  e-prints}, p. arXiv:1511.00363, Nov. 2015.

\bibitem{KnowlDistil15}
G.~{Hinton}, O.~{Vinyals}, and J.~{Dean}, ``{Distilling the Knowledge in a
  Neural Network},'' \emph{arXiv e-prints}, p. arXiv:1503.02531, Mar. 2015.

\bibitem{LinStruct14}
E.~{Denton}, W.~{Zaremba}, J.~{Bruna}, Y.~{LeCun}, and R.~{Fergus},
  ``{Exploiting Linear Structure Within Convolutional Networks for Efficient
  Evaluation},'' \emph{arXiv e-prints}, p. arXiv:1404.0736, Apr. 2014.

\bibitem{PruningReed93}
R.~Reed, ``Pruning algorithms-a survey,'' \emph{IEEE Transactions on Neural
  Networks}, vol.~4, no.~5, pp. 740--747, 1993.

\bibitem{SurveyCompressHardware}
L.~Deng, G.~Li, S.~Han, L.~Shi, and Y.~Xie, ``Model compression and hardware
  acceleration for neural networks: A comprehensive survey,'' \emph{Proceedings
  of the IEEE}, vol. 108, no.~4, pp. 485--532, 2020.

\bibitem{Bengio13_Whitney16}
Y.~Bengio, A.~Courville, and P.~Vincent, ``Representation learning: A review
  and new perspectives,'' \emph{IEEE Transactions on Pattern Analysis and
  Machine Intelligence}, vol.~35, no.~8, pp. 1798--1828, 2013.

\bibitem{Goodfellow-et-al-2016}
I.~Goodfellow, Y.~Bengio, and A.~Courville, \emph{Deep Learning}.\hskip 1em
  plus 0.5em minus 0.4em\relax MIT Press, 2016,
  \url{http://www.deeplearningbook.org}.

\bibitem{Higgins2016betaVAELB}
I.~Higgins, L.~Matthey, A.~Pal, C.~P. Burgess, X.~Glorot, M.~M. Botvinick,
  S.~Mohamed, and A.~Lerchner, ``beta-vae: Learning basic visual concepts with
  a constrained variational framework,'' in \emph{ICLR}, 2016.

\bibitem{dsprites17}
L.~Matthey, I.~Higgins, D.~Hassabis, and A.~Lerchner, ``dsprites:
  Disentanglement testing sprites dataset,''
  https://github.com/deepmind/dsprites-dataset/, 2017.

\bibitem{chen2018isolating}
R.~T. Chen, X.~Li, R.~B. Grosse, and D.~K. Duvenaud, ``Isolating sources of
  disentanglement in variational autoencoders,'' \emph{NeurIPS}, vol.~31, 2018.

\end{thebibliography}
\end{document}